\documentclass{article}
\pdfoutput=1

\PassOptionsToPackage{numbers, compress}{natbib}

\usepackage[preprint]{neurips_2023}

\usepackage[utf8]{inputenc} 
\usepackage[T1]{fontenc}    
\usepackage{hyperref}       
\usepackage{url}            
\usepackage{booktabs}       
\usepackage{amsfonts}       
\usepackage{nicefrac}       
\usepackage{microtype}      
\usepackage{xcolor}         

\usepackage{graphicx}
\usepackage{algorithm, algorithmic}
\usepackage{threeparttable}

\DeclareUnicodeCharacter{0308}{\"{u}}

\title{Molecular Property Prediction Based on Graph Structure Learning}

\author{
  Bangyi Zhao \\
  Fudan University\\
  Shanghai\\
  \texttt{byzhao21@m.fudan.edu.cn} \\
  \And
  Weixia Xu \\
  Fudan University \\
  Shanghai \\
  \texttt{xuweixia@fudan.edu.cn} \\
  \AND
  Jihong Guan \\
  Tongji University \\
  Shanghai \\
  \texttt{jhguan@tongji.edu.cn} \\
  \And
  Shuigeng Zhou\thanks{Corresponding author} \\
  Fudan University \\
  Shanghai \\
  \texttt{sgzhou@fudan.edu.cn} \\
}

\begin{document}

\maketitle

\begin{abstract}
Molecular property prediction (MPP) is a fundamental but challenging task in the computer-aided drug discovery process. More and more recent works employ different graph-based models for MPP, which have made considerable progress in improving prediction performance. However, current models often ignore relationships between molecules, which could be also helpful for MPP. For this sake, in this paper we propose a graph structure learning (GSL) based MPP approach,
called GSL-MPP. Specifically, we first apply graph neural network (GNN) over molecular graphs to extract molecular
representations. Then, with molecular fingerprints, we construct a molecular similarity graph (MSG). Following that, we
conduct graph structure learning on the MSG (i.e., molecule-level graph structure learning) to get the final molecular
embeddings, which are the results of fusing both GNN encoded molecular representations and the relationships among
molecules, i.e., combining both intra-molecule and inter-molecule information. Finally, we use these molecular embeddings
to perform MPP. Extensive experiments on seven various benchmark datasets show that our method could achieve state-of-the-art performance in most cases, especially on classification tasks. Further visualization studies also demonstrate the
good molecular representations of our method.
\end{abstract}

\section{Introduction}
The accurate prediction of molecular properties is a critical task in the field of drug discovery. By utilizing computational methods, this task can be accomplished with great efficiency, reducing both time and expense associated with identifying drug candidates. This is particularly important considering that the average cost of developing a new drug is currently estimated to be approximately \$2.8 billion~\cite{fleming2018artificial,wieder2020compact} and the development period lasts a dozen of years, let alone the high risk of clinical failure~\cite{sarkar2023artificial}. Naturally, a molecule can be abstracted as a topological graph, where atoms are treated as nodes and bonds are viewed as edges. In the past few years, deep graph learning methods, especially various graph neural networks (GNNs) have been applied in this field, offering effective molecular graph representations for accurate molecular property prediction~\cite{duvenaud2015convolutional,song2020communicative,sun2020graph}. In GNNs, nodes iteratively update their representations after aggregating information from their neighbours and a final graph-pooling layer will generate a graph representation for the molecule. Up to now, various message passing layers have been proposed and applied, including GAT~\cite{velivckovic2017graph}, MPNN~\cite{gilmer2017neural} and GIN~\cite{xu2018powerful}. And later studies further considered to integrate edge features into the passing messages in order to improve the expressive power of their models, like DMPNN~\cite{yang2019analyzing} and CMPNN~\cite{song2020communicative}.

Despite the considerable progress, most of recent studies focus only on the message passing within individual molecules. The relationships among molecules are often ignored, which could also play an important role in property prediction~\cite{wang2021hierarchical}. One relatively easy and effective way is to construct a relationship graph among molecules using the structural similarity, because a critical assumption of medicinal chemistry is that structurally similar molecules tend to have similar biological activities~\cite{johnson1990concepts}. For example, fingerprint (carrying the structural information of the molecules) similarity search is often used in virtual screening~\cite{muegge2016overview}. However, this assumption is not always true since a phenomenon called activity cliff (AC) exits. An AC is defined as a pair of structurally similar compounds with a large potency difference against a given target~\cite{maggiora2006outliers,stumpfe2012exploring,stumpfe2014recent,stumpfe2020advances}. Thus, the relationship graph constructed by structural similarity may be not ``perfect'' for the downstream tasks. We need to take certain measures to enhance this relationship graph if we want to make full and proper use of it.

To address these problems above, we propose a novel two-level graph representation learning method for molecular property prediction, called GSL-MPP. Our method operates in a two-level molecular graph representation framework: (i) Atom-level molecular graph representation where molecular graphs composed of atoms and bonds represent the intra-structures of molecules; and (ii) molecule-level graph representation where inter-molecule similarity graph (MSG in short) is constructed by fingerprint similarity to encode similarities between molecules that allows effective label propagation among connected similar molecules. Intra-molecular representation is done by GNNs, and inter-molecular representation is finished by graph structure learning (GSL). This two-level graph representation enables us to comprehensively exploit both intra-molecule and inter-molecule information to get better molecular representations and overcome (to some degree) the AC problem, consequently boosting MPP performance.

Specifically, we applies metric-based iterative graph structure learning in our method. The MSG structure and molecular embeddings are updated for $T$ times. During each iteration, GSL-MPP learns a better MSG structure based on better molecular embeddings, and in turn, learns better molecular embeddings with a better MSG structure. Besides, during the training process, we also add a GSL-specific loss to the common supervised loss for better MSG structure learning on both classification tasks and regression tasks. Our method is evaluated on 7 benchmark datasets including 4 classification tasks and 3 regression tasks.  Experimental results show that our model can achieve state-of-the-art performance in most cases. Ablation studies show that the combination of fingerprint similarity and GSL is of particular effectiveness.
\section{Related Work}
\subsection{Molecular Property Prediction}
Most methods for predicting molecular properties can be summarized using a general framework. In this framework, we first transform the input molecule $m$ into a specific-length vector $h$ using a representation function, $h = g(m)$. Then another prediction function is used to predict a specific property $y$ based on $h$, $y = f(h)$. During this period, a good molecular representation is of vital importance to address molecular property prediction problems.

At early stages, traditional chemical fingerprints such as Extended Connectivity Fingerprints (ECFP)~\cite{Morgan1965MorganAlgorithm,glen2006circular} are used to encode a molecule to a vector. These fingerprints could carry the structural information of the molecules~\cite{nguyen2023perceiver}.

In order to improve the expressive power, recent works started to use the graph neural networks (GNNs) to acquire graph-level representation as molecular embedding. Examples include graph convolutional network (GCN)~\cite{duvenaud2015convolutional}, graph attention network (GAT)~\cite{velivckovic2017graph}, message passing neural network (MPNN)~\cite{gilmer2017neural} and graph isomorphism network (GIN)~\cite{xu2018powerful}.Later works extend the MPNN framework to consider bond information during message passing procedure, like DMPNN~\cite{yang2019analyzing} and CMPNN~\cite{song2020communicative}. Besides, CD-MVGNN~\cite{ma2022cross} also considers both atom-level and bond-level message passing, and a cross-dependency mechanism is designed to ensure these two views rely on information from each other during feature updates, thereby enhancing expressive capabilities.

Recently, many efforts have also been made to integrate transformer to graph neural network. Molecule Attention Transformer(MAT)~\cite{maziarka2020molecule} attempts to  incorporate node distance and graph structural information when calculating attention scores. Another work Grover~\cite{rong2020self} combines message-passing networks with the Transformer architecture to create a more expressive molecular encoder that captures information at two hierarchical levels. CoMPT~\cite{chen2021learning} is also built upon the Transformer architecture. Unlike previous graph Transformer models that treated molecules as fully connected graphs, this approach employs a message diffusion mechanism inspired by heat diffusion phenomena to integrate information from the adjacency matrix, alleviating the over-smoothing issue.

However, these methods only focus on the structure of a single molecule, while ignoring the important role of inter-molecular relationships for property prediction.
\subsection{Graph Structure Learning}
The expressive power of GNNs often depends on the input graph structure. However, the initial graph structure is not always optimal for downstream tasks. On the one hand, the original graph is constructed from the original feature space, which may not reflect the "true" graph topology after feature extraction and transformation. On the other hand, errors can also occur when data is measured or collected, making the graph noisy or even incomplete. Graph structure learning (GSL) is one of the methods that can effectively solve this problem, through learning and optimizing the graph structure~\cite{zhu2021survey}. Recently, ~\cite{chen2020iterative} proposed the method of iterative deep graph learning (IDGL) for jointly and iteratively learning graph structure and node embeddings in the field of natural language processing (NLP). 
It was later used by ~\cite{wang2021property} for few-shot molecular property prediction. Compared to ~\cite{wang2021property}, our method is not based on few-shot situation and the datasets and baselines we choose are not for few-shot either. Besides, The specific implementation of GSL is different. More importantly, we try to construct an initial graph between molecules before we apply GSL, which is confirmed to be necessary in ablation study.

\section{Method}\label{sec:method}
\subsection{Overview}
\begin{figure}
    \centering
    \includegraphics[width=1.0\textwidth]{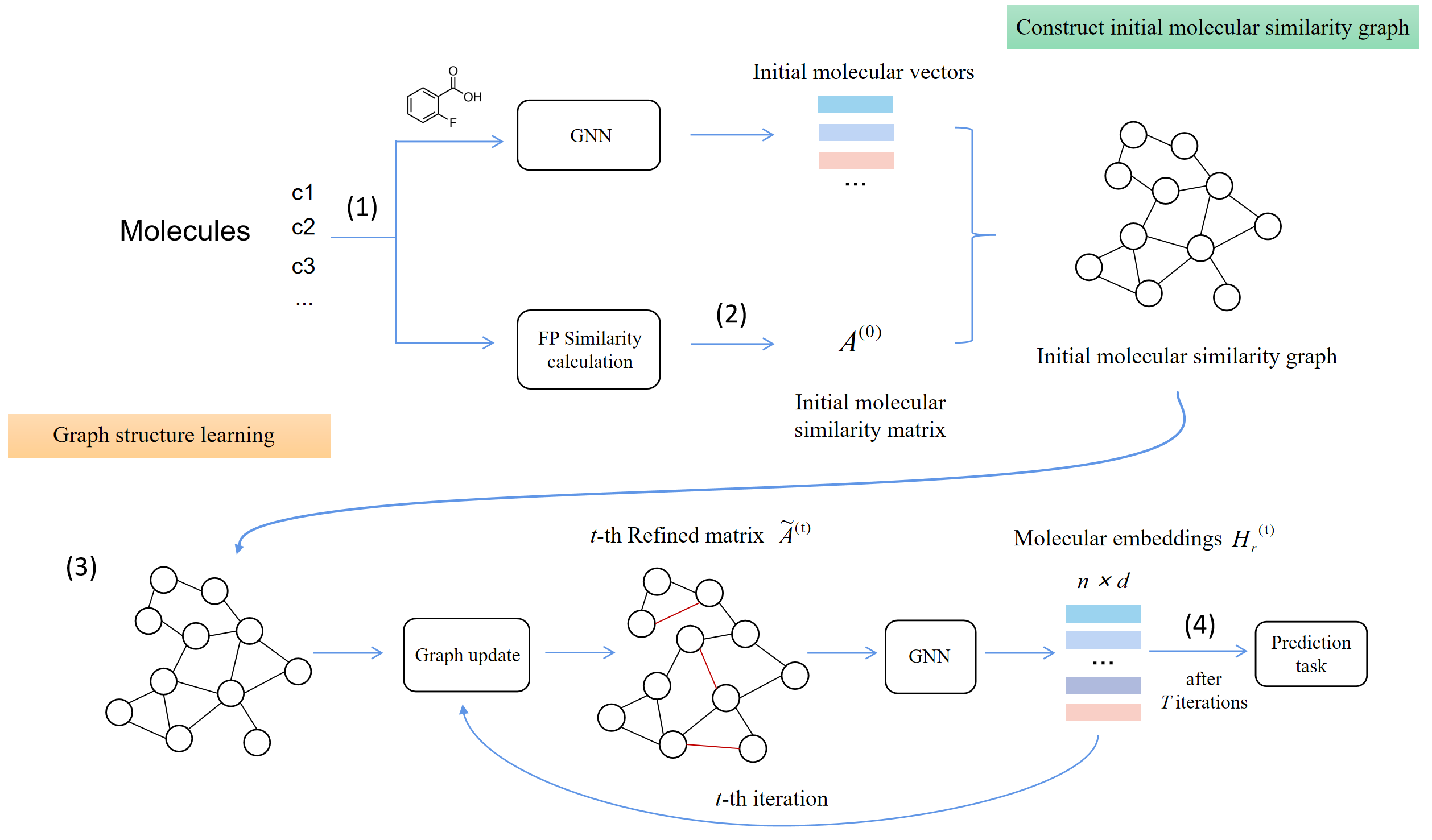}
    \caption{The workflow of GSL-MPP. In the initial molecular similarity graph (MSG), each node is a molecule initially represented by the GNN and each edge is attached with the FP similarity between the two corresponding molecules. GSL is then performed on the MSG.}
    \label{fig:model}
\end{figure}

The structure of our method GSL-MPP is illustrated in Fig.~\ref{fig:model}, which is operated on a two-level graph learning framework. Specifically, the two-level graph learning framework consists of (i) the lower level: atom-level molecular graphs encoded by GNN to extract the initial molecular representations, and (ii) the upper level: a molecule-level similarity graph, on which graph structure learning (GSL) is performed to iteratively learn the final molecular embeddings, where inter-molecular relationships are exploited. 

The workflow of GSL-MPP is as follows: (1) molecule graphs are first encoded by a GNN to obtain initial molecular embeddings. Meanwhile, molecules are represented as feature vectors using fingerprints. (2) With the molecular feature vectors, the initial molecular similarity graph (MSG) is constructed, where each node is a molecule  initially represented by the above GNN embeddings, and each edge attached with a weight --- the similarity between the two corresponding molecules. (3) GSL is performed on the MSG, which 
 iteratively updates the molecular embeddings and the graph structure. (4) The final molecular embeddings are used for property prediction. 

\subsection{Molecular Graph Embedding}
Here, we describe how to represent a molecular graph as an initial vector by GNN. 
A molecule $m$ can be abstracted as an attributed graph where $G_m = (\mathcal{V}, \mathcal{E})$, in which $\left| \mathcal{V}\right| = n_v$ refers to a set of $n_v$ atoms (nodes) and $\left| \mathcal{E}\right| = n_e$ refers to a set of $n_e$ bonds (edges) in the molecule. $x_v$ are used to represent the initial feature of node $v$ and $\mathcal{N}_v$ denotes the set of neighbors of node $v$.

\subsubsection{Node embedding}
We use Graph Isomorphism Network (GIN)~\cite{xu2018powerful} as intra-molecule GNN to extract each node's embedding:
\begin{equation}
    h_v^{(k)} = MLP^{(k)} ((1 + \epsilon^{(k)}) \cdot h_v^{(k-1)} + \sum_{u\in \mathcal{N}(v)} h_u^{(k-1)}),
\end{equation}
where MLP means multi-layer perceptron, $h_v^{(k)}$ is the representation vector of node $v$ at the $k$-th layer. We initialize $h_v^{(0)} = x_v$, and $\epsilon$ is a learnable parameter.

\subsubsection{Graph pooling}
After gaining each node's embedding, a READOUT operation is applied to get the initial molecular embedding $h_g$:
\begin{equation}
    h_g = READOUT(\left\{ \left. h_v^{k}\right| v \in G \right\} \left| k = 0, 1, ..., K \right.).
\end{equation}

\subsection{Constructing Molecular Similarity Graph (MSG)}
Our inter-molecule graph reflects the relationships between molecules, where each node indicates a molecule, and each edge means the relationship between two molecules. As shown in Fig.~\ref{fig:model}, the initial feature vector of each node is the molecule's embedding obtained by GNN (their embedding matrix is denoted as $X_r$), and the initial adjacency matrix $A^{(0)}$ is calculated by the structural similarity between molecules. Here, we calculate the structural similarity between molecules based on their Extended Connectivity Fingerprints (ECFP)~\cite{Morgan1965MorganAlgorithm}. 

ECFPs are circular fingerprints, possessing several beneficial characteristics: 1) They can be calculated fast; 2) They are not predefined and can capture an almost limitless range of molecular characteristics including stereochemical information; 3) They indicate the presence of specific substructures, facilitating interpretation of computation results~\cite{rogers2010extended}.
Specifically, we get each molecule's ECFP and calculate the Tanimoto Coefficient as the similarity score. A hyperparameter $\epsilon_{tc}$ acts as a threshold to obtain a sparse matrix. That is, we mask off those elements in the adjacency matrix that are smaller than $\epsilon_{tc}$. We apply molecular fingerprints to construct $A^{(0)}$ because it contains useful structural information~\cite{nguyen2023perceiver} and could offer an informative initial inter-molecule graph.

\subsection{Structure Learning on Molecular Similarity Graph}
As we have discussed, this similarity graph constructed above may not be good enough for downstream tasks, therefore here graph structure learning is employed to enhance the graph by exploiting inter-molecule relationships. Specifically, initial matrix built with fingerprint similarity only measure structural similarity between molecules and may not ``perfectly'' reflect true molecular property similarity, so we use GSL to refine it. The core of GSL is the structure learner that could be grouped into three types: (1) Metric-based approaches use a metric function like cosine similarity on pairwise node embeddings to calculate edge weights; (2) Neural approaches employ neural networks to infer edge weights; and (3) Direct approaches treat all elements of the adjacency matrix as learnable parameters~\cite{zhu2021survey}.

In this paper, following IDGL~\cite{chen2020iterative}, we adopt the metric-based approach and employ $m$-perspective weighted cosine similarity as the metric function:
\begin{equation}
    s_{ij}^p = \cos (w_p \odot v_i, w_p \odot v_j), \quad s_{ij} = \frac{1}{m} \sum_{p=1}^{m} s_{ij}^p,
\label{eq:similarity-metric} 
\end{equation}
where $s_{ij}^p$ estimates the cosine similarity between nodes $v_i$ and $v_j$, each perspective $p$ considers one part of the semantics contained in the vectors and corresponds to a learnable weight vector $w_p$. The obtained $s_{ij}$ is the entry in row $i$ and column $j$ of the newly learned adjacency matrix $A$. Also the $\epsilon$-neighborhood sparsification technique is applied to obtaining a sparse and non-negative adjacency matrix.    

The node embeddings $H_r$ and the adjacency matrix $A$ will be alternately refined for $T$ times. At the $t$-th iteration, $A^{(t)}$ is calculated from the previously updated node embeddings $H_r^{(t-1)}$ by Eq.~(\ref{eq:similarity-metric}). Then we use the learned graph structure $A^{(t)}$ as supplementary to optimize the initial graph $A^{(0)}$:
\begin{equation}
    \widetilde A^{(t)} = \lambda A^{(0)} + (1 - \lambda)\left\{ \eta A^{(t)} + (1 - \eta) A^{(1)}   \right\},
\label{eq:update-adj} 
\end{equation}
where $A^{(1)}$ is the adjacency matrix learned from $X_r$ at the 1-st iteration in order to maintain the initial node information. $\lambda$ and $\eta$ are hyperparameters.

After learning the adjacency matrix, we employ an $L$-layer inter-molecule GNN to learn node embeddings, and in the $l$-th layer, $H_r^{(t, l)}$ is updated by
\begin{equation}
    H_r^{(t, l)} = ReLU (\widetilde A^{(t)} H_r^{(t, l-1)} W_r^{(l)}),
\label{eq:update-embeddings} 
\end{equation}
$H_r^{(t)} = H_r^{(t, L)}$ is the final node embeddings in this iteration and $H_r^{(t, 0)} = X_r$. 

\subsection{Loss Function}
After $T$ rounds of iteration, the node (molecule) embeddings $H_r^{(T)}$ represent the final molecular representations. Based on this, predictions can be made for specific property $\hat y$ with a fully connected layer (FC) as follows:
\begin{equation}
    \hat y = FC(H_r^{(T)}).
\label{eq:prediction} 
\end{equation}

The whole loss function used in our method consists of two parts: the label prediction loss and the GSL-specific loss. The label prediction loss function $\mathcal{L}_{pred}$ is obtained in a manner similar to existing methods:
\begin{equation}
    \quad \mathcal{L}_{pred} = \ell (\hat y, y),
    \label{eq:prediction-loss} 
\end{equation}
where $\hat y$ represents the predicted value, $y$ is the ground truth, and $\ell$ represents the loss function used. In classification tasks, it is the Cross Entropy Loss, and in regression tasks, it is the Mean Squared Error Loss.

Since the quality of the learned inter-molecule graph structure is of great importance for our method, we further design a GSL-specific loss, hoping that the learned adjacency matrix does not contain wrong edges.
We use $S_{train}$ to represent molecules in training set and $\widetilde A^{(T)}$ to represent the final adjacency matrix after being refined $T$ times. In classification tasks, there exists a ground truth for the matrix, $A^*$($A^*_{ij} = 1$ if $y_i = y_j$ else 0), i.e., molecules with the same label should be connected by edges. Thus, we define the GSL-specific loss as
\begin{equation}
    \mathcal{L}_{GSL} = \sum_{x_i, x_j \in S_{train}} (\widetilde A_{ij}^{(T)} - A_{ij}^*)^2
    \label{eq:gsl-loss-clf} 
\end{equation}
However, in regression tasks, the prediction of a molecule is a real value and no native ground truth exists. We have to define it by ourselves. For the convenience of calculation, we only consider those molecular pairs with large difference (beyond a certain threshold $\epsilon_y$) in predicted values when calculating the GSL-specific loss:
\begin{equation}
    \mathcal{L}_{GSL} = \sum_{x_i, x_j \in S_{train}} (\widetilde A^{(T)}_{ij})^2 , \quad  x_i, x_j \quad satisfy \quad \left| y_i - y_j \right| > \epsilon_y.
    \label{eq:gsl-loss-reg}
\end{equation}
The whole loss function combines both the task prediction loss and the GSL-specific loss, that is, $\mathcal{L} = \mathcal{L}_{pred} + \mathcal{L}_{GSL}$.

\subsection{Algorithm}
The algorithm of our method is presented in Algorithm 1. After obtaining the initial molecular embeddings and constructing the initial inter-molecule similarity graph MSG (corresponding to the adjacency matrix), $T$ iterations of GSL are applied. During each iteration, the adjacency matrix is refined based on the node embeddings gained in the last iteration,  while the node embeddings are updated based on adjacency matrix obtained in the last iteartion.

\begin{algorithm}
\caption{The GSL-MPP algorithm}
\begin{algorithmic}[1]
\STATE Obtain initial molecular embedding $h_{g, i}$ for each molecule $m_i$ by a graph-based molecular encoder (an intra-molecule GNN);
\STATE $X_r \leftarrow$ embedding matrix of all $h_{g, i}$;
\STATE $H_r^{(0)} \leftarrow X_r$;
\STATE Construct an initial molecule similarity matrix $A^{(0)}$ using molecular fingerprint similarity;
\FOR {$t = 1$ to $T$}          
    \STATE Use GSL to learn a refined adjacency matrix $A^{(t)}$ by $H_r^{(t - 1)}$ using Eq.~(\ref{eq:similarity-metric});
    \STATE Combine initial and refined adjacency matrices $A^{(0)}$ and $A^{(t)}, A^{(1)}$ to obtain $\widetilde A^{(t)}$ by Eq.~(\ref{eq:update-adj});
    \STATE Initialize node embeddings by $H_r^{(t, 0)} = X_r$;
    \FOR {$l = 1$ to $L$}
        \STATE Update node embedding $H_r^{(t, l)}$ by inter-molecule GNN using Eq.~(\ref{eq:update-embeddings});
    \ENDFOR
    \STATE $H_r^{(t)} \leftarrow H_r^{(t, L)}$;
\ENDFOR
\STATE Obtain prediction $\hat y$ using $H_r^{(T)}$ by Eq.~(\ref{eq:prediction});
\IF{in training phase}
\STATE Calculate $\mathcal{L}_{pred}$ by Eq.~(\ref{eq:prediction-loss}) and $\mathcal{L}_{GSL}$ by Eq.~(\ref{eq:gsl-loss-clf}) or Eq.~(\ref{eq:gsl-loss-reg}) for $S_{train}$;
\STATE $\mathcal{L} \leftarrow \mathcal{L}_{pred} + \mathcal{L}_{GSL}$;
\STATE Back-propagate $\mathcal{L}$ to update model weights;
\ENDIF
\end{algorithmic}
\end{algorithm}

\section{Performance Evaluation}\label{sec:performance}

\subsection{Experimental Setting}
\paragraph{Datasets.}We use 7 benchmark datasets from MoleculeNet~\cite{wu2018moleculenet} for experiments, among which 4 are classification tasks and 3 are regression
tasks. Specifically, BACE is about the binding results of several inhibitors; BBBP is the blood–brain barrier penetration dataset;
SIDER and Clintox are two multi-task datasets corresponding to side effects and toxicity respectively; ESOL, Lipophilicity and Freesolv are regression datasets about physical chemistry properties.

Scaffold splitting of ~\cite{yang2019analyzing} is adopted to split the
datasets into training, validation, and test, with a 0.8/0.1/0.1
ratio, which is more empirical and challenging than random splitting. Following previous works~\cite{ma2022cross,rong2020self}, we use three independent runs on three random-seeded scaffold splitting for each dataset.

\paragraph{Baselines.} We compare our method against 12 baselines. 
TF\_Robust~\cite{ramsundar2015massively} is a DNN-based multitask framework that takes molecular fingerprints
as input. GCN~ (GraphConv)~\cite{duvenaud2015convolutional}, Weave~\cite{kearnes2016molecular} and SchNet~\cite{schutt2017schnet} are three graph convolutional models. MPNN~\cite{gilmer2017neural} and its variants MGCN~\cite{lu2019molecular}, DMPNN~\cite{yang2019analyzing} and CMPNN~\cite{song2020communicative} are models considering the edge features during message passing. AttentiveFP~\cite{xiong2019pushing} is an extension of the graph attention network. GROVER~\cite{rong2020self} and CoMPT~\cite{chen2021learning} are two transformer-based models. Here, GROVER is compared  without the pretrain process for a fair comparison. CoMPT is a transformer-based model utilizing both nodes and edges information in message passing process while CD-MVGNN~\cite{ma2022cross} also constructs two views for atoms and bonds respectively.

\paragraph{Evaluation metrics.} Following the evaluation criteria adopted by these baseline models, we use AUC-ROC to evaluate the performance of classification tasks, and RMSE to evaluate regression tasks.

\paragraph{Implementation details.} Our model apply a polynomial decay scheduler to the learning rate with two linear increase warm-up epochs and polynomial decay afterward. The power of polynomial decay is set to 1, indicating a linear decay. The final learning rate is 1e-9 and the max\_epoch is 300. For the proposed model, on each dataset we try different hyper-parameter combinations, and take the hyper-parameter set with the best result. While building the initial inter-molecule graph, the radius of used ECFP is 2. The threshold of GSL-specific loss for regression tasks ($\epsilon_y$) is 0.01. More details of the hyper-parameter setting in the implementation of our model are presented in Table \ref{table:hyper-parameter}. 

\begin{table}
\caption{Hyper-parameter settings.}
\label{table:hyper-parameter}
\resizebox{\columnwidth}{!}{
\begin{tabular}{ccc}
\hline
Hyper-parameter    & Description                                                            & Value range                        \\ \hline
max\_lr            & maximum learning rate of  polynomial decay scheduler                   & 0.0001$\sim$0.01             \\
weight\_decay      & weight\_decay weight decay percentage for Adam optimizer               & 0.00001$\sim$0.001           \\
gin\_layers        & number of the intra-molecule GIN layers                                & 2$\sim$5                     \\
gin\_hidden\_size  & number of the hidden dimensionality in the intra-molecule GIN          & 32, 64, 128, 256             \\
tc\_epsilon        & threshold of Tanimoto Coefficient for $A^{(0)}$ ($\epsilon_{tc}$)                               & 0.0, 0.1, 0.2, 0.3, 0.5, 0.7 \\
gsl\_iter          & number of the iterations for graph structure learning ($T$)              & 1$\sim$5                     \\
gsl\_gnn\_layers   & number of the inter-molecule GCN layers                                & 2, 3                         \\
gsl\_hidden\_size  & number of the hidden dimensionality in the inter-molecule GCN          & 32, 64, 128, 256             \\
gsl\_epsilon       & threshold of similarity score in GSL ($\epsilon_{gsl}$)                                    & 0, 0,1, 0,2, 0.5             \\
gsl\_perspective   & number of perspective used in GSL ($m$)                                      & 1, 2, 4, 8, 16               \\
gsl\_skip\_conn    & the ratio of initial matrix while updating the graph structure ($\lambda$)      & 0.1, 0.3, 0.5, 0.7, 0.9      \\
gsl\_update\_ratio & the ratio of t-th learned matrix while updating the graph structure ($\eta$) & 0.1, 0.3, 0.6, 0.8. 1.0      \\
dropout            & dropout rate                                                           & 0, 0.1, 0.2, 0.4, 0.6        \\ 
gsl\_coff          & the coefficient of the GSL related loss ($\mu$)                                & 0.1, 0.3, 0.5, 0.7, 0.9 \\

\hline
\end{tabular}
}
\end{table}

\begin{table}
\caption{Performance comparison between our model and baselines. Mean and standard deviation of AUC or RMSE values are reported.}
\label{table:experiment}
\renewcommand{\arraystretch}{1.5}
\resizebox{\columnwidth}{!}{\begin{tabular}{c|cccc|ccc}
\hline
            & \multicolumn{4}{c|}{Classifcation (ROC-AUC)}                                                                                                              & \multicolumn{3}{c}{Regression (RMSE)}                                                                         \\ \hline
Dataset     & \multicolumn{1}{c|}{BACE}                 & \multicolumn{1}{c|}{BBBP}                 & \multicolumn{1}{c|}{ClinTox}              & SIDER                & \multicolumn{1}{c|}{FreeSolv}             & \multicolumn{1}{c|}{ESOL}                 & Lipop                \\ \hline
TFRobust    & \multicolumn{1}{c|}{0.824±0.022}          & \multicolumn{1}{c|}{0.860±0.087}          & \multicolumn{1}{c|}{0.765±0.085}          & 0.607±0.033          & \multicolumn{1}{c|}{4.122±0.085}          & \multicolumn{1}{c|}{1.722±0.038}          & 0.909±0.060          \\
GraphConv   & \multicolumn{1}{c|}{0.854±0.011}          & \multicolumn{1}{c|}{0.877±0.036}          & \multicolumn{1}{c|}{0.845±0.051}          & 0.593±0.035          & \multicolumn{1}{c|}{2.900±0.135}          & \multicolumn{1}{c|}{1.068±0.050}          & 0.712±0.049          \\
Weave       & \multicolumn{1}{c|}{0.791±0.008}          & \multicolumn{1}{c|}{0.837±0.065}          & \multicolumn{1}{c|}{0.823±0.023}          & 0.543±0.034          & \multicolumn{1}{c|}{2.398±0.250}          & \multicolumn{1}{c|}{1.158±0.055}          & 0.813±0.042          \\
SchNet      & \multicolumn{1}{c|}{0.750±0.033}          & \multicolumn{1}{c|}{0.847±0.024}          & \multicolumn{1}{c|}{0.717±0.042}          & 0.545±0.038          & \multicolumn{1}{c|}{3.215±0.755}          & \multicolumn{1}{c|}{1.045±0.064}          & 0.909±0.098          \\
MPNN        & \multicolumn{1}{c|}{0.815±0.044}          & \multicolumn{1}{c|}{0.913±0.041}          & \multicolumn{1}{c|}{0.879±0.054}          & 0.595±0.030          & \multicolumn{1}{c|}{2.185±0.952}          & \multicolumn{1}{c|}{1.167±0.430}          & 0.672±0.051          \\
DMPNN       & \multicolumn{1}{c|}{0.852±0.053}          & \multicolumn{1}{c|}{0.919±0.030}          & \multicolumn{1}{c|}{0.897±0.040}          & 0.632±0.023          & \multicolumn{1}{c|}{2.177±0.914}          & \multicolumn{1}{c|}{0.980±0.258}          & 0.653±0.046          \\
MGCN        & \multicolumn{1}{c|}{0.734±0.030}          & \multicolumn{1}{c|}{0.850±0.064}          & \multicolumn{1}{c|}{0.634±0.042}          & 0.552±0.018          & \multicolumn{1}{c|}{3.349±0.097}          & \multicolumn{1}{c|}{1.266±0.147}          & 1.113±0.041          \\
CMPNN       & \multicolumn{1}{c|}{0.869±0.023}          & \multicolumn{1}{c|}{0.929±0.025}          & \multicolumn{1}{c|}{0.922±0.017}          & 0.617±0.016          & \multicolumn{1}{c|}{2.060±0.505}          & \multicolumn{1}{c|}{0.838±0.140}          & \underline{0.625±0.027}    \\
AttentiveFP & \multicolumn{1}{c|}{0.863±0.015}          & \multicolumn{1}{c|}{0.908±0.050}          & \multicolumn{1}{c|}{0.933±0.020}          & 0.605±0.060          & \multicolumn{1}{c|}{2.030±0.420}          & \multicolumn{1}{c|}{0.853±0.060}          & 0.650±0.030          \\
CD-MVGNN    & \multicolumn{1}{c|}{\textbf{0.892±0.011}} & \multicolumn{1}{c|}{\underline{0.933±0.006}}    & \multicolumn{1}{c|}{\underline{0.945±0.037}}    & \underline{0.639±0.012}    & \multicolumn{1}{c|}{\textbf{1.552±0.123}} & \multicolumn{1}{c|}{\textbf{0.779±0.026}} & \textbf{0.553±0.013} \\ \hline
GROVER\textsuperscript{*}     & \multicolumn{1}{c|}{0.858}                & \multicolumn{1}{c|}{0.911}                & \multicolumn{1}{c|}{0.884}                & 0.624                & \multicolumn{1}{c|}{1.987}                & \multicolumn{1}{c|}{0.911}                & 0.643                \\
CoMPT       & \multicolumn{1}{c|}{0.838±0.035}          & \multicolumn{1}{c|}{0.926±0.028}          & \multicolumn{1}{c|}{0.876±0.031}          & 0.612±0.026          & \multicolumn{1}{c|}{2.006±0.628}          & \multicolumn{1}{c|}{0.822±0.090}          & 0.663±0.035          \\ \hline
OurModel    & \multicolumn{1}{c|}{\underline{0.871±0.038}}    & \multicolumn{1}{c|}{\textbf{0.957±0.008}} & \multicolumn{1}{c|}{\textbf{0.947±0.020}} & \textbf{0.652±0.014} & \multicolumn{1}{c|}{\underline{1.974±0.315}}    & \multicolumn{1}{c|}{\underline{0.799±0.118}}    & 0.693±0.063          \\ \hline
\end{tabular}}
\begin{tablenotes}
	\item *Here, GROVER does not use pretrained model for a fair comparison, and standard deviation is not provided in its the original paper.
     \end{tablenotes}
\end{table}
\subsection{Performance Comparison}
Table \ref{table:experiment} presents the results of our model and the baselines on all datasets. boldfaced values are the best results, and underlined values are the 2nd best results. From  \ref{table:experiment} we have the following observations: (1) Compared to the SOTA model CD-MVGNN, Our model performs better on 3/4 classification datasets with a 2.4\% AUC lift on BBBP. Since our model is based on a simple GIN without complicated message passing procedures used in CD-MVGNN and CoMPT, this result indicates the effectiveness of the inter-molecule graph for prediction tasks. (2) Our model performs relatively poor on regression tasks compared to SOTAs, which may be explained by the lack of real ground truth of relationship graphs in regression tasks. However, our model still achieve 2nd best results on 2/3 datasets. (3) Though our model uses GIN for intra-molecule graphs, it outperforms GIN on 7 of the 8 datasets. Especially, on the ClinTox dataset, our model gets up to 7.8\% performance improvement. This also reflects the effectiveness of our molecule similarity graph construction and graph structure learning.

\subsection{Ablation Study}
To investigate the contribution of each component of our model, an ablation study is conducted. We consider four variant models for comparison as follows:
\begin{itemize}
    \item \textbf{Not any}: directly use $H_r^{(0)}$ to predict. It is almost a GIN network.
    \item \textbf{Only $A^{(0)}$}: apply GNN on the initial molecular similarity graph $A^{(0)}$ constructed by ECFP similarity without GSL.
    \item \textbf{Only GSL}: use de novo GSL without an initial graph reference $A^{(0)}$.
    \item \textbf{No GSL-Loss}: use $A^{(0)}$ and GSL, but apply only the prediction loss.
\end{itemize}

Ablation results are given in Table \ref{table:ablation}. Here we mainly consider the contribution of the initial adjacency matrix $A^{(0)}$ constructed by ECFP fingerprints and GSL process in our method. The results of ``Not any'', ``Only $A^{(0)}$'' and ``No GSL-Loss'' confirm that the use of $A^{(0)}$ could improve the performance of our model and the improvement will be much more significant when combined with GSL. Besides, it is interesting to notice that ``Only GSL'' often performs worse than ``Not any", which probably means learning an inter-molecule graph from scratch might be difficult and it is necessary for us to utilize the chemical information of molecular fingerprints to build an initial graph. Finally, while comparing ``No GSL-Loss" and the complete ``Our Model", we can see that GSL-specific loss does make a difference for our method.

We also conduct experiments to show the results of using different values of some important hyper-parameters on all the datasets. 
Table \ref{table:lambda} reports the results of applying different values of $\lambda$, which is used to balance the learned graph structure and the initial graph structure. It can be seen that applying a large $\lambda$ value (0.8 or 0.9) will generate a relatively good results on most datasets, which indicates the importance of the initial inter-molecule graph.   
Besides, Table \ref{table:T} shows the impact of the number of iteration $T$ on performance. We can see that as $T$ increases from 1 to 5,  performance on most datasets does not show  continuous improvement, which means that the best $T$ is data dependent. 

\begin{table}
\caption{Ablation study on four variants of our model.}
\label{table:ablation}
\renewcommand{\arraystretch}{1.5}
\resizebox{\columnwidth}{!}{\begin{tabular}{c|cccc|ccc}
\hline
            & \multicolumn{4}{c|}{Classification (ROC-AUC)}                                                                                                             & \multicolumn{3}{c}{Regression (RMSE)}                                                                         \\ \hline
     & \multicolumn{1}{c|}{BACE}                 & \multicolumn{1}{c|}{BBBP}                 & \multicolumn{1}{c|}{ClinTox}              & SIDER                & \multicolumn{1}{c|}{FreeSolv}             & \multicolumn{1}{c|}{ESOL}                 & Lipop                \\ \hline
Not any      & \multicolumn{1}{c|}{0.809±0.032}          & \multicolumn{1}{c|}{0.943±0.005}          & \multicolumn{1}{c|}{0.932±0.012}          & 0.593±0.017          & \multicolumn{1}{c|}{2.055±0.405}          & \multicolumn{1}{c|}{0.962±0.093}          & 0.888±0.046          \\
only A0     & \multicolumn{1}{c|}{0.856±0.049}          & \multicolumn{1}{c|}{0.951±0.002}          & \multicolumn{1}{c|}{0.939±0.033}          & 0.639±0.030          & \multicolumn{1}{c|}{3.433±0.552}          & \multicolumn{1}{c|}{0.945±0.076}          & 0.724±0.040          \\
only GSL    & \multicolumn{1}{c|}{0.850±0.025}          & \multicolumn{1}{c|}{0.943±0.017}          & \multicolumn{1}{c|}{0.875±0.049}          & 0.580±0.021          & \multicolumn{1}{c|}{2.638±0.098}          & \multicolumn{1}{c|}{1.147±0.256}          & 0.899±0.196          \\
No GSL-Loss & \multicolumn{1}{c|}{0.865±0.034}          & \multicolumn{1}{c|}{0.953±0.015}          & \multicolumn{1}{c|}{0.935±0.016}          & 0.651±0.026          & \multicolumn{1}{c|}{2.134±0.155}          & \multicolumn{1}{c|}{0.821±0.100}          & 0.711±0.047          \\
Our Model   & \multicolumn{1}{c|}{\textbf{0.871±0.038}} & \multicolumn{1}{c|}{\textbf{0.957±0.008}} & \multicolumn{1}{c|}{\textbf{0.947±0.020}} & \textbf{0.652±0.014} & \multicolumn{1}{c|}{\textbf{1.974±0.315}} & \multicolumn{1}{c|}{\textbf{0.799±0.118}} & \textbf{0.693±0.063} \\ \hline
\end{tabular}}
\end{table}

\begin{table}
\caption{Results for different $\lambda$ values on different datasets.}
\label{table:lambda}
\renewcommand{\arraystretch}{1.5}
\resizebox{\columnwidth}{!}{\begin{tabular}{cccccccc}
\hline
lambda & BACE                 & BBBP                 & ClinTox              & SIDER                & FreeSolv             & ESOL                 & Lipop                \\ \hline
0.1    & 0.702±0.1            & 0.924±0.006          & 0.914+0.026          & 0.564+0.035          & 2.812+0.239          & 0.983+0.094          & 0.714+0.056          \\
0.3    & 0.846±0.062          & 0.944±0.008          & \textbf{0.947±0.020} & 0.606+0.018          & 2.371+0.171          & 0.986+0.112          & 0.749+0.05           \\
0.5    & 0.839±0.043          & 0.921±0.017          & 0.939+0.018          & 0.616+0.007          & 2.224+0.231          & 0.87+0.147           & 0.706+0.059          \\
0.7    & 0.849±0.049          & 0.942±0.004          & 0.916+0.037          & 0.634+0.006          & 2.126+0.153          & 0.931+0.056          & 0.725+0.066          \\
0.8    & 0.850±0.028          & \textbf{0.957±0.008} & 0.927+0.023          & \textbf{0.652±0.014} & \textbf{1.974±0.315} & 0.827+0.09           & \textbf{0.693±0.063} \\
0.9    & \textbf{0.871±0.038} & 0.939±0.013          & 0.925+0.023          & 0.643+0.004          & 2.279+0.089          & \textbf{0.799±0.118} & 0.707+0.066          \\ \hline
\end{tabular}
}
\end{table}

\begin{table}
\caption{Results for  different $T$ values  on different datasets.}
\label{table:T}
\renewcommand{\arraystretch}{1.5}
\resizebox{\columnwidth}{!}{
\begin{tabular}{cccccccc}
\hline
$T$ & BACE                 & BBBP                 & ClinTox              & SIDER                & FreeSolv             & ESOL                 & Lipop                \\ \hline
1 & 0.848±0.062          & 0.945±0.007          & 0.93±0.028           & 0.64±0.017           & 2.724±0.484          & 0.863±0.046          & 0.718±0.047          \\
2 & 0.858±0.045          & 0.923±0.023          & \textbf{0.947±0.020} & 0.636±0.015          & \textbf{1.974±0.315} & 0.86±0.127           & 0.751±0.053          \\
3 & 0.857±0.043          & 0.944±0.008          & 0.93±0.005           & \textbf{0.652±0.014} & 2.182±0.364          & 0.872±0.087          & \textbf{0.693±0.063} \\
4 & \textbf{0.871±0.038} & \textbf{0.957±0.008} & 0.907±0.018          & 0.64±0.007           & 2.209±0.41           & 0.861±0.093          & 0.752±0.047          \\
5 & 0.851±0.057          & 0.937±0.011          & 0.918±0.034          & 0.63±0.015           & 2.309±0.291          & \textbf{0.799±0.118} & 0.751±0.058          \\ \hline
\end{tabular}
}
\end{table}

\begin{figure}
    \centering
    \includegraphics[width=1.0\textwidth]{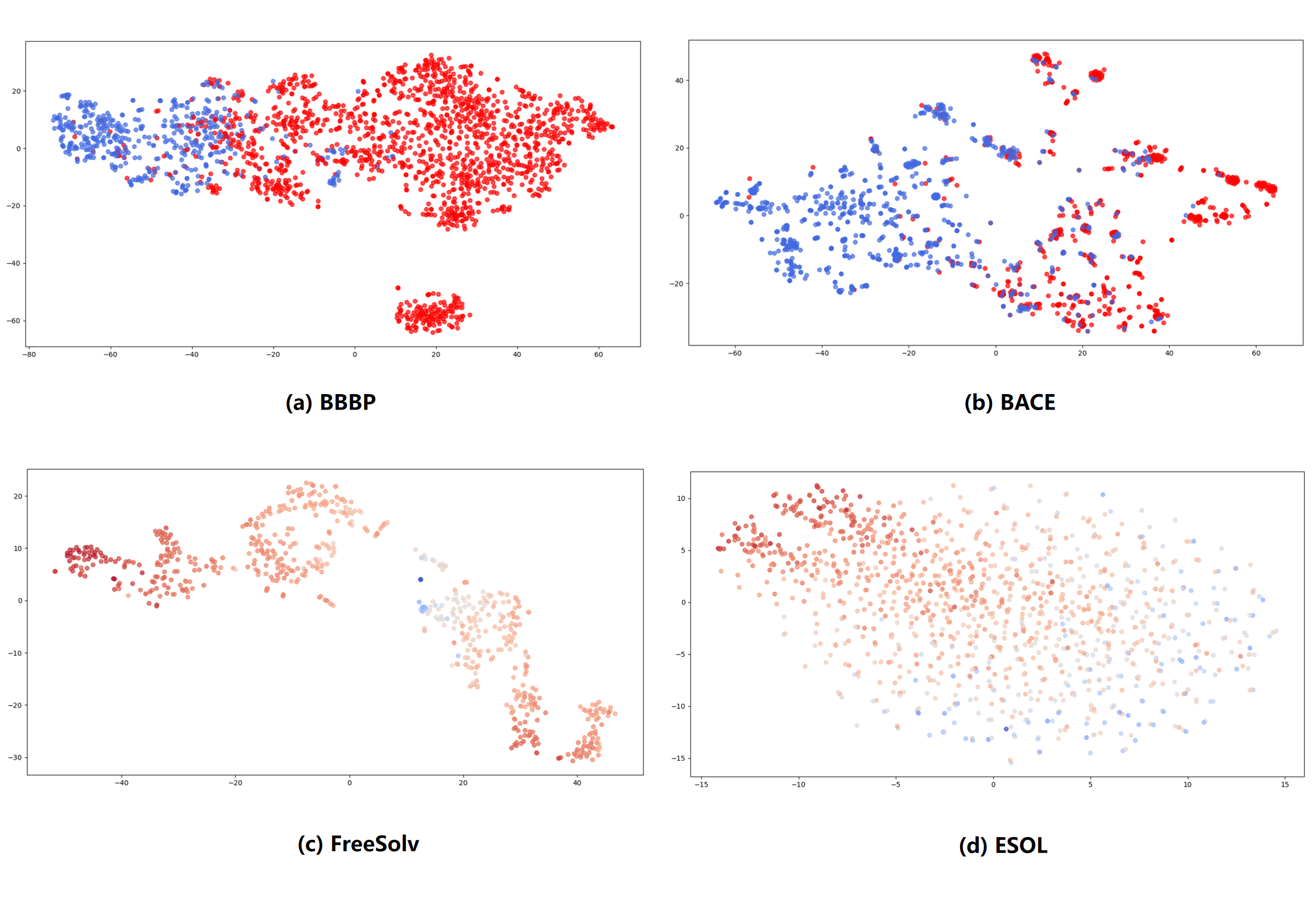}
    \caption{Visualization of molecular representations for 4 datsets: (a) BBBP,  (b) BACE,  (c) FreeSolv and (d) ESOL. For classification datasets BACE and BBBP, molecules of label 1 are colored in red and molecules of label 0 are colored in blue. For regression datasets FreeSolv and ESOL, the colors of the points change from red to blue as the property value increases.}
    \label{fig:tsne}
\end{figure}

\subsection{Visualization of Molecular Representations}
To check the molecular representation learning ability of our model, we apply t-distributed Stochastic Neighbor Embedding (t-SNE) with default hyper-parameters to visualize the final molecular representations on four datasets, including two classification datasets (BACE and BBBP) and two regression datasets (FreeSolv and ESOL). The results are shown in Fig.~\ref{fig:tsne}.

We can see that molecules of different labels have a clear boundary for both two classification datasets, especially for BBBP. Molecules of the same label tend to be clustered together, while molecules of different labels are located apart. Also, there seems a certain distribution pattern existing among the molecules of different property values for the two regression datasets. For the FreeSolv dataset, molecules tend to move from the outer region to the inner region as the property value decreases. As for the ESOL dataset, molecules tend to move from upper left to lower right as the property value decreases. These results indicate that our model generates reasonable representations of molecules for downstream tasks.

\subsection{Scaling Our Model to Larger Datasets}
During GSL, the similarity metric function calculate similarity scores for all pairs of graph nodes, which requires $\mathcal{O}(n^2)$ complexity. So we need to address the scalability issue if the size of datasets becomes larger. Following IDGL~\cite{chen2020iterative}, we apply an anchor-based method. During each iteration, We learn a node-anchor similarity matrix $R \in \mathbb{R}^{n \times s}$ instead of the original complete adjacency matrix $A \in \mathbb{R}^{n \times n}$. $s$ represents the number of anchor nodes, which is a hyperparameter that can be set according to different datasets. By using $R$ instead of $A$, the time and space complexity can be reduced from $\mathcal{O}(n^2)$ to $\mathcal{O}(ns)$.  Therefore, Eq. (3) in the paper can be rewritten as the following:
\begin{equation}
    s_{ik}^p = \cos (w^p \odot v_i, w^p \odot u_k), \quad s_{ik} = \frac{1}{m} \sum_{p=1}^{m} s_{ik}^p
\end{equation}
where $s_{ik}$ is the similarity score between node $v_i$ and anchor $u_k$. The procedure of message passing should also be changed accordingly. The node-anchor similarity matrix $R$ allows only direct connections between nodes and anchors. We call a direct travel between a node and an anchor as one-step transition described by $R$. Based on theories of stationary Markov random walks, we can actually recover $A$ from $R$ by calculating the two-step transition probabilities.

Using the above anchor-based GSL, We firstly evaluate whether introducing anchor nodes will have a great impact on the original prediction performance of our model. Results are given in Table \ref{table:anchor-baseline}. We can find that anchor-based GSL performs a little worse than the original GSL in these molecule datasets but the performance degradation is not significant. So we think it is appropriate for us to apply anchor-based GSL in larger-scale molecule datasets.

\begin{table}
\caption{Performance comparison between original and anchor-based GSL.}
\label{table:anchor-baseline}
\resizebox{\columnwidth}{!}{
\begin{tabular}{c|c|c|c|c|c|c|c}
\hline
             & BACE  & BBBP  & ClinTox & SIDER & FreeSolv & ESOL  & Lipop \\ \hline
Origin       & 0.865 & 0.953 & 0.935   & 0.651 & 2.134    & 0.821 & 0.711 \\ \hline
Anchor-based & 0.818 & 0.949 & 0.95    & 0.612 & 2.208    & 0.794 & 0.747 \\ \hline
\end{tabular}
}
\end{table}

\begin{table}
\centering
\caption{Performance comparison between our model (using anchor-based GSL) and baselines.}
\label{table:anchor-hiv}
\begin{tabular}{c|c}
\hline
            & ROC-AUC\%       \\ \hline
Our model   & \textbf{81.8} \\
PharmHGT    & 80.6          \\
DMPNN       & 78.6          \\
CD-MVGNN    & 78.4          \\
CoMPT       & 78.1          \\
CMPNN       & 77.4          \\
AttentiveFP & 75.7          \\
MPNN        & 74.1          \\
GROVER      & 62.5          \\ \hline
\end{tabular}
\end{table}

After completing the above evaluation, we test the anchor-based GSL method on the HIV dataset which includes over 40000 molecules and compare it with some existing models. Except for CD-MVGNN, the results of other models on the HIV data set are from PharmHGT~\cite{jiang2023pharmacophoric}. PharmHGT is a recently proposed model based on the Transformer structure, which treats molecules as heterogeneous graphs. The ROC-AUC of CD-MVGNN on the HIV data set is obtained experimentally by ourselves. Results are given in Table \ref{table:anchor-hiv}. Our method is able to achieve the optimal ROC-AUC on the HIV dataset, showing that after introducing anchor nodes, our method can be well extended to larger-scale datasets and achieve satisfactory results.

\section{Conclusion}\label{sec:conclusion}
In this paper, we propose a new model based on two-level molecular representation for molecular property prediction. Unlike previous attempts focusing exclusively on message passing between atoms or bonds within individual molecule graphs, we further take use of the inter-molecule graph. Concretely, we utilize the chemical information of molecular fingerprints to construct an initial molecular similarity graph, and employ graph structure learning to refine the graph. Molecular embeddings based on GSL on the inter-molecular similarity graph are used for MPP. 
Extensive experiments show that our model can achieve state-of-the-art performance in most cases, especially on the classification tasks. Ablation studies also validate the major designed components of the model. 

However, there is still room to improve our model in the following directions: (1) Using more sophisticated graph-based models to encode molecular graphs rather than GIN. (2) Designing new metrics other than weighted cosine similarity for graph structure learning. (3) Exploring new and more effective GSL methods.

\bibliographystyle{plain}
\bibliography{reference}

\end{document}